\documentclass[conference]{IEEEtran}
\IEEEoverridecommandlockouts
\usepackage{mathtools}
\usepackage{balance}
\usepackage{subcaption}
\usepackage{tikz}
\usepackage{multirow}
\usepackage{caption}
\usepackage{makecell}

\usepackage{cite}
\usepackage{amsmath,amssymb,amsfonts}
\usepackage{algorithmic}
\usepackage{graphicx}
\usepackage{textcomp}
\usepackage{xcolor}
\def\BibTeX{{\rm B\kern-.05em{\sc i\kern-.025em b}\kern-.08em
    T\kern-.1667em\lower.7ex\hbox{E}\kern-.125emX}}
\begin{document}

\title{Graph Attention Recurrent Neural Networks for Correlated Time Series Forecasting---Full Version}

\author{
\IEEEauthorblockN{Razvan-Gabriel Cirstea}
\IEEEauthorblockA{\textit{Department of Computer Science} \\
\textit{Aalborg University}\\
Aalborg, Denmark \\
razvan@cs.aau.dk}
\and

\IEEEauthorblockN{Chenjuan Guo}
\IEEEauthorblockA{\textit{Department of Computer Science} \\
\textit{Aalborg University}\\
Aalborg, Denmark \\
cguo@cs.aau.dk}
\and

\IEEEauthorblockN{Bin Yang}
\IEEEauthorblockA{\textit{Department of Computer Science} \\
\textit{Aalborg University}\\
Aalborg, Denmark \\
byang@cs.aau.dk}
\and
}

\maketitle

\begin{abstract}We consider a setting where multiple entities interact with each other over time and the time-varying statuses of the entities are represented as multiple \emph{correlated time series}. 
For example, speed sensors are deployed in different locations in a road network, where the speed of a specific location across time is captured by the corresponding sensor as a time series, resulting in multiple speed time series from different locations, which are often correlated.  
To enable accurate forecasting on correlated time series, we proposes \emph{graph attention recurrent neural networks}. 
First, we build a graph among different entities by taking into account spatial proximity and employ a multi-head attention mechanism to derive adaptive weight matrices for the graph to capture the correlations among vertices (e.g., speeds at different locations) at different timestamps.  
Second, we employ recurrent neural networks to take into account temporal dependency while taking into account the adaptive weight matrices learned from the first step to consider the correlations among time series. 
Experiments on a large real-world speed time series data set suggest that the proposed method
is effective and outperforms the state-of-the-art in most settings. This manuscript provides a full version of a workshop paper~\cite{MileTS}.
\end{abstract}


\section{Introduction}

\par 

Complex cyber-physical systems (CPSs) often consist of multiple entities that interact with each other across time. 
With the continued digitization, various sensor technologies are deployed to record time-varying
attributes of such entities, thus producing correlated time series~\cite{DBLP:conf/cikm/CirsteaMMG018,DBLP:journals/pvldb/0002GJ13}.
One representative example of such CPSs is road transportation system~\cite{DBLP:journals/vldb/YangDGJH18,DBLP:journals/tc/Ding0C016}, where the speeds on different roads are captured by different sensors, such as loop detectors, speed cameras, and bluetooth speedometers, as multiple speed time series~\cite{DBLP:journals/vldb/HuYGJ18,DBLP:journals/sigmod/GuoJ014,DBLP:journals/geoinformatica/HuYJM17}. 
%
%

Accurate forecasting of correlated time series have the potential to reveal holistic system dynamics of the underlying CPSs, including 
predicting future behavior~\cite{DBLP:conf/cikm/CirsteaMMG018} and detecting anomalies~\cite{DBLP:conf/mdm/Kieu0J18,IJCAI19}, which are important to enable effective operations of the CPSs.  
For example, in an intelligent transportation system, analyzing speed time series enables travel
time forecasting, early warning of congestion, and predicting the effect of incidents, which help drivers make routing decisions~\cite{DBLP:journals/tkde/LiuJYZ18,DBLP:conf/icde/Guo0HJ18}. 

\begin{figure}[!h]
    \centering
    \includegraphics[width=\columnwidth]{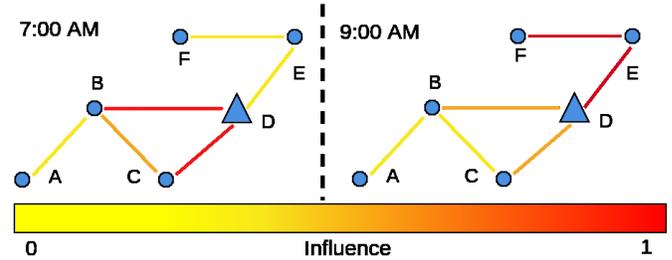}
    \caption{Dynamic Interactions}
    \vspace{-15pt}
    \label{fig:motivation}
\end{figure}

Traffic on different roads often has an impact of the neighbour roads. An accident on one road might cause congestion on the others. Most of the related work capture such interactions as a graph which is pre-computed based on different metrics such as road or euclidean distance between pairwise entities. 
Although such methods are easy to implement, they only capture static interactions between entities. In contrast, the interactions among entities are often dynamic and evolve across time.
Figure \ref{fig:motivation} shows a graph of 6 vertices representing 6 entities, e.g., speed sensors. At 7 AM vertex $D$ has strong connections with vertices $B$ and $C$, while at 9 AM, the connections with vertices $B$ and $C$ become weaker but vertex $D$ has a stronger connection with vertex $E$ instead. 
However, most of the recent studies fall short on capturing such dynamics~\cite{DCRNN,STGCN,MRGNN,DBLP:journals/pvldb/0002GJ13}.

To enable accurate and robust correlated time series forecasting, it is essential to model such spatio-temporal correlations among multiple time series. To this end, we propose \emph{graph attention recurrent neural networks (GA-RNNs)}. 

We first build a graph among different entities by taking into account spatial proximity. In the graph, vertices represent entities and two vertices are connected by an edge if the two corresponding entities are nearby. 
After building the graph, we apply multi-head attention to learn an attention  matrix. For each vertex, the attention matrix indicates, among all the vertex's neighbor vertices, which neighboring vertices' speeds are more relevant when predicting the speed of the vertex. 
%

Next, since recurrent neural networks (RNNs) are able to well capture temporal dependency, we modify classic RNNs to capture spatio-temporal dependency. Specifically, we replace weight multiplications in classic RNNs with convolutions that take into account graph topology (e.g., graph convolution or diffusion convolution~\cite{DCRNN}). However, instead of using a static adjacency matrix, we employ the attention matrix learned from the first step to obtain adaptive adjacency matrices. Here, with the learned attention weight matrices and the inputs at different timestamps, we obtain different adjacency matrices at different timestamps, which are able to capture the dynamic correlations among different time series at different timestamps. 

The main contribution is to utilize attention to derive adaptive adjacency matrices which are then utilized in RNNs to capture spatio-temporal correlations among time series to enable accurate forecasting for correlated time series. 
The proposed graph attention is a generic approach which can be applied to any graph-based convolution operations that utilizes adjacency matrices such as graph convolution and diffusion convolution~\cite{DCRNN}. 
Experiments on a large real-word traffic time series offer evidence that the proposed method is accurate and robust. 
\section{Problem definition}

We first consider the temporal aspects by introducing multiple time series and then the spatial aspects by defining graph signals. Finally, we define correlated time series forecasting.

\subsection{Time Series}

Consider a cyber-physical system (CPS) where we have $N$ entities. 
The status of the $i$-th entity across time, e.g., from timestamps 1 to $T$, is represented by a time series $TS^{(i)}=\langle \mathbf{x}_1^{(i)}, \mathbf{x}_2^{(i)}, \ldots, \mathbf{x}_T^{(i)} \rangle$, where a $K$-dimensional vector $\mathbf{x}_j^{(i)}$ records $K$ features (e.g., speed and traffic flow) of the $i$-th entity at timestamp $j$. 

Since we have a time series for each entity, we have in total $N$ time series: $TS^{(1)}$, $TS^{(2)}$, \ldots, $TS^{(N)}$. 

Given historical statuses of all entities, we aim at predicting the future statuses of all entities. 
More specifically, from the time series, we have historical statuses covering
a window $[t_{a-L+1}, t_{a}]$ that contains $L$ time stamps, and we aim
at predicting the future statuses in a future window
$[t_{a+1}, t_{a+P}]$ that contains $P$ time stamps. 
We call this problem \emph{$p$-step ahead forecasting}. 

\subsection{Graph Signals}

We still consider the CPS with $N$ entities. Now, we focus on the modeling of interactions among these entities.  

We build a directed graph $G=(V, E)$ where each vertex $v\in V$ represents an entity in the CPS, which is often associated with spatial information such as longitude and latitude. Since we have $N$ entities in total, we have $|V|=N$. Edges are represented by adjacency matrix $E\in \mathbb{R}^{N \times N}$, where $E[i, j]=1$ represents an edge from the i-th to the j-th vertices. 

If the entities are already embedded into a spatial network, e.g., camera sensors deployed in road intersections in a road network, we connect two entities by an edge if they are connected in the spatial network. Otherwise, we connect two entities by an edge if the distance between them  
is smaller then a threshold, which is shown to be effective~\cite{DCRNN}. 

\begin{figure}[ht]
    \centering
    \begin{subfigure}[b]{0.4\columnwidth} 
        \centering
        \includegraphics[width=\columnwidth]{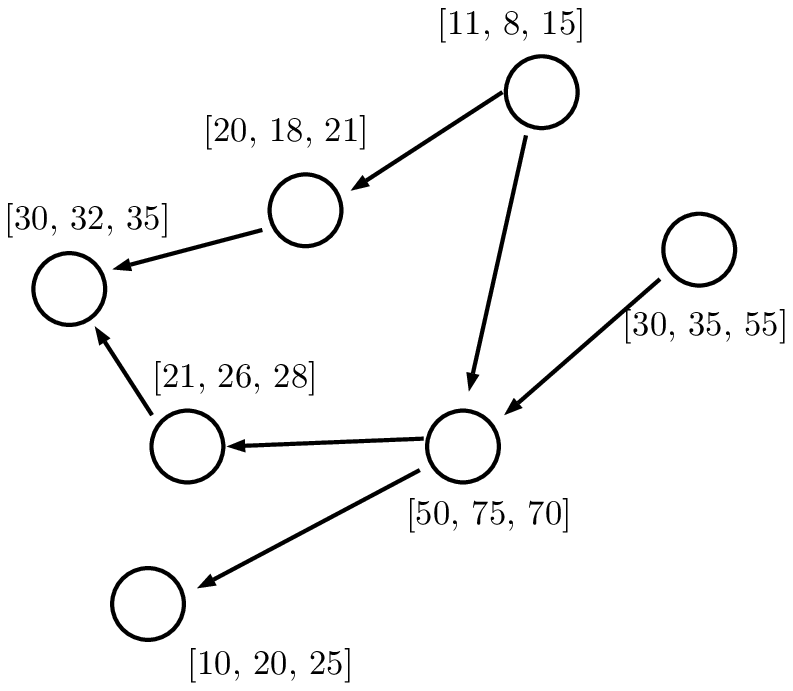}
        \caption{Graph signal at a timestamp}
        \label{fig:graph_signal}
    \end{subfigure}
    \hfill
    \begin{subfigure}[b]{0.53\columnwidth}
        \centering
        \includegraphics[width=\columnwidth]{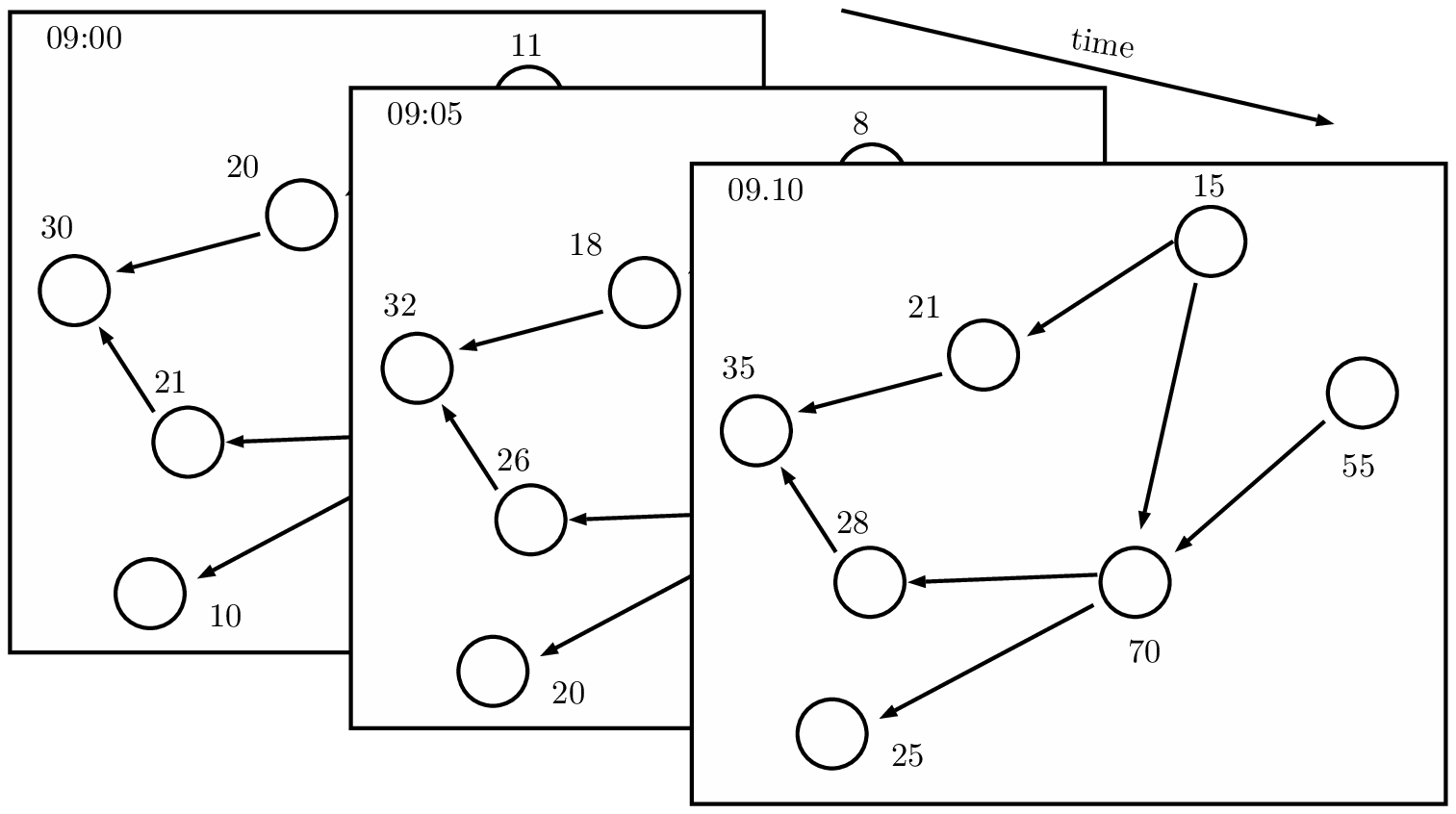}
        \caption{Graph signals at different timestamps}
        \label{fig:graph_signal_at_time}
    \end{subfigure}
    \caption{A Time Series of Graph Signals, $K=3$.}
    \label{fig:signals}
\end{figure}

At each timestamp $t$, each entity is associated with $K$ features (e.g., speed and traffic flow). We introduce a graph signal at $t$, as shown in Figure \ref{fig:graph_signal}, where $\mathbf{X}_t \in \mathbb{R}^{N \times K}=[\mathbf{x}_t^{(1)}, \mathbf{x}_t^{(2)}, \ldots, \mathbf{x}_t^{(N)}]^T$ represent all features from all entities at timestamp $t$. 
Based on the concept of graph signals, the problem becomes learning a function that takes as input $L$ past graph signals and outputs $P$ future graph signals: 
$$
\mathbf{X}^{(L)} \rightarrow \mathbf{X}^{(P)}, 
$$
where $\mathbf{X}^{(L)}=[\mathbf{X}_{(t-L+1)}, \ldots, \mathbf{X}_{(t-1)}, \mathbf{X}_{(t)}]$, $\mathbf{X}^{(P)}=[\mathbf{X}_{(t+1)}, \mathbf{X}_{(t+2)}, \ldots, \mathbf{X}_{(t+P)}]$, and $t$ is the current timestamp.

%



\section{Graph Attention RNNs}

We proceed to describe the proposed graph attention recurrent neural network (\emph{GARNN}) to solve the $p$-step ahead forecasting. 




\subsection{GARNN Framework}

\begin{figure*}
\resizebox {2\columnwidth} {!}{
    \input{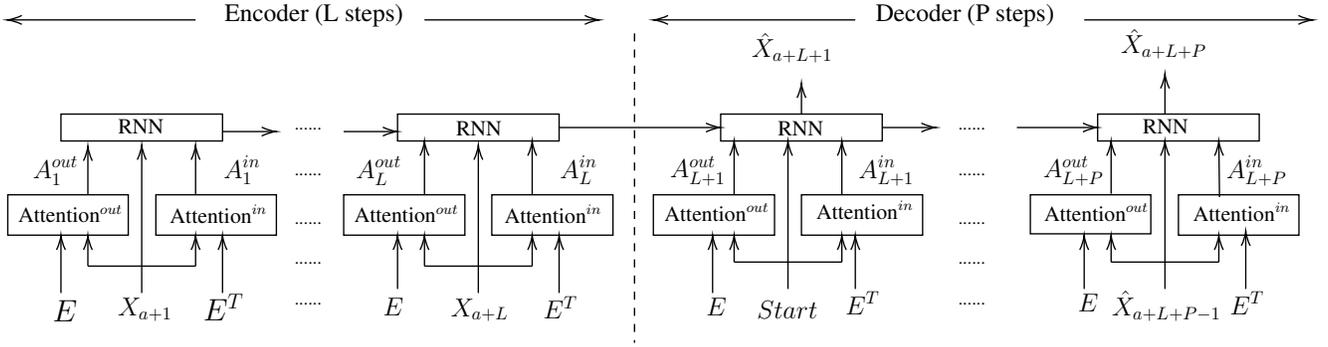}
}
\caption{Graph Attention Recurrent Neural Network}
\vspace{-15pt}
\label{fig:encoder_decoder}
\end{figure*}

\par The proposed GARNN consists of two parts, an attention part and an RNN part, which follow an encoder-decoder architecture as shown in Fig.~\ref{fig:encoder_decoder}. 
At each timestamp, we firstly model the spatial correlations among different entities using multi-head attention as two adjacency matrices $\mathbf{A}_t^{out}$ and $\mathbf{A}_t^{in}$, which consider the outgoing and incoming traffic, respectively. Next, the two attention weight matrices are fed into an RNN unit together with the input graph signal at the time stamp, which facilitate the RNN unit not only capture the temporal dependency but also the spatial correlations, making it is possible to capture the spatio-temporal correlations among different time series.  


\subsection{Spatial Modeling}

\par To capture the spatial correlations among different entities at a specific timestamp,  we employ attention mechanism~\cite{attention}. 
The idea is to determine, in order to predict the features of an entity, i.e., a vertex, how much should we consider the features of the vertex's neighbour vertices. We may consider different neighbour vertices differently at different timestamps. 
Specifically, for each vertex $i$, we compute an attention score w.r.t. to each vertex in $NB(i)=\{j|E[i,j]=1\}\cup \{i\}$, i.e., vertex $i$'s neighbour vertices and itself. 

\par We proceed to show how to compute an attention score for vertex $i$. 
Recall that for any vertex $i$ in the graph, its features at timestamp $t$ is represented by a $K$-dimensional vector $\mathbf{x}_t^{(i)}\in \mathbb{R}^{K}$. Vertex $j$ is a vertex from $NB(i)$. 
Attention score $\mathbf{A}_t[i, j]$ indicates how much attention should be paid to vertex $j$'s features in order to estimate vertex $i$'s features at timestamp $t$, which is computed based on Equation~\ref{eq:attention_head}, where $[\cdot||\cdot]$ represents the concatenation operator.%

\begin{equation}
   \small
    A_t[i,j] = \frac{exp(LReLU(\mathbf{v}^\intercal [\mathbf{W}\mathbf{x}_t^{(i)} \; || \; \mathbf{W}\mathbf{x}_t^{(j)}]))}{\sum_{m \in NB(i)} exp(LReLU(\mathbf{v}^\intercal [\mathbf{W}\mathbf{x}_t^{(i)} \; || \; \mathbf{W}\mathbf{x}_t^{(m)}]))}
\label{eq:attention_head}
\end{equation}

First, we embed the features from each vertex with an embedding matrix $\mathbf{W} \in \mathcal{R}^{F \times K}$, where $F$ is the size of the embedding. 
Then, we concatenate the embeddings of the two vertices' features $\mathbf{Wx}_t^{(i)}$ and $\mathbf{Wx}_t^{(j)}$, which is fed into an attention network. 
The attention network is constructed as a single-layer feed forward neural network, parameterized by a weight matrix $\mathbf{v} \in \mathbb{R}^{2\cdot F\times 1}$. In addition, we apply a LeakyReLU function $\mathit{LReLU}$ with a negative input slope of $a=0.2$ as activation function (see Euqation ~\ref{eq:leaky}). Finally, we apply softmax to normalize the final output to obtain $\mathbf{A}_t[i, j]$ so that the results are easier to interpret.

\begin{equation}
    \mathit{LReLU}(x) = 
    \begin{dcases}
    ax,& \text{if } x\leq 0\\
    x,              & \text{if } x \; \textgreater \; 0
    \end{dcases}
    \label{eq:leaky}
\end{equation}

\par Motivated by~\cite{attention_all_you_need}, we observe that stacking multiple attention networks, a.k.a., attention heads, is beneficial since each attention network can specialise on capturing different interactions. 
Assume that we use a total of $C$ attention networks, each network has its own embedding matrix $\mathbf{W}^{(c)}\in \mathcal{R}^{F \times K}$ and weight matrix $\mathbf{v}^{(c)}\in \mathbb{R}^{2\cdot F\times 1}$ in the feed forward neural network. 
There are multiple ways of combining the attention heads, in our experiments we use average according to Equation~\ref{multihead_att}, 

\begingroup\abovedisplayskip=3pt \belowdisplayskip=2pt
\begin{equation}
\small
    \mathbf{A}_t[i,j] = \frac{1}{C}\sum_{c=1}^C \frac{exp(LReLU(\mathbf{v}_{c}^\intercal [\mathbf{W}_{c} \mathbf{x}_t^{(i)}  \; || \; \mathbf{W}_{c} \mathbf{x}_t^{(j)} ]))}{\sum_{m \in NB(i)} exp(LReLU(v_{c}^\intercal [\mathbf{W}_{c} \mathbf{x}_t^{(i)}  \; || \; \mathbf{W}_{c} \mathbf{x}_t^{(j)} ]))}
\label{multihead_att}
\end{equation}
\endgroup

\par So far, for each vertex, we have computed the attentions to all its neighbors, which captures the influence of the outgoing traffic. On the other hand, it is also of interest to capture the influence of the incoming traffic. To this end, we use the transpose of the adjacency matrix $E$ to define neighboring vertices and apply the same attention mechanism to obtain another attention matrix to represent the influence of incoming traffic. 
Finally, we obtain two attention matrices $\mathbf{A}_t^{out}\in \mathbb{R}^{N \times N}$ and $\mathbf{A}_t^{in}\in \mathbb{R}^{N \times N}$, which capture the influence from outgoing traffic and incoming traffic, respectively.  

\subsection{Spatio-Temporal Modeling}

We integrate the learned attention matrices into classical recurrent neural networks to capture spatio-temporal correlations.  
We follow an encoder-decoder architecture as shown in Figure~\ref{fig:encoder_decoder}, which is able to well capture the temporal dependencies across time. 
Next, we replace the matrix multiplications in RNN units by convolutions that take into account graph topology such as diffusion convolution or graph convolution. 
Here, instead of using a static adjacency matrix $E$ that only captures connectivity, the convolution here employs the learned adjacency matrices $\mathbf{A}_t^{out}$ and $\mathbf{A}_t^{out}$ at timestamp $t$. 
Note that although the attention embedding and weight matrices $\mathbf{W}_c$ and $\mathbf{v}_c$ are static across time, the input matrices $\mathbf{x}_t^{(i)}$ are changing across time. Thus, we are able to derive different adjacency matrices $\mathbf{A}_t^{out}$ and $\mathbf{A}_t^{out}$ at different timestamps. 

More specifically, we proceed to define a diffusion convolution operator $\otimes$ on a graph signal $\mathbf{X}_t \in \mathcal{R}^{N \times K}$ using the learned attention matrices $\mathbf{A}_t^{out}$ and $\mathbf{A}_t^{in}$ at timestamp $t$ in Equation~\ref{eq:diffusion}.  

\begingroup\abovedisplayskip=0pt \belowdisplayskip=2pt
\begin{equation}
\small
\begin{split}
X_t \otimes \Theta = \alpha (\sum_{k=1}^{K} \sum_{h=1}^{H} (\theta_{k, h, 1} (\mathbf{A}_t^{out})^h + &\theta_{k, h, 2}  (\mathbf{A}_t^{in})^h)  \mathbf{X}_{t}[\cdot, k])
\label{eq:diffusion}
\end{split}
\end{equation}
\endgroup

%
%
Here, $\alpha$ is an activation function, matrix $\Theta \in \mathcal{R}^{K\times H \times 2}$ is a filter to be learned, and $\mathbf{X}_{t}[\cdot, k]$ represents the $k$-th column of graph signal $\mathbf{X}_t$, which is the $k$-th features of all entities.  
We often apply $Q$ different filters to perform diffusion convolutions and then concatenate the results into a matrix $\mathbf{X}_t \otimes \mathbf{\Theta}=[\mathbf{X}_t \otimes \Theta_1 || \mathbf{X}_t \otimes \Theta_2|| \ldots|| \mathbf{X}_t \otimes \Theta_Q]$. Finally, graph signal $\mathbf{X}_t$ is convoluted into matrix $\mathbf{X}_t \otimes \mathbf{\Theta} \in \mathcal{R}^{N\times Q}$.  


%

\par Next, we integrate the proposed graph attention based convolution into an RNN. 
Here, we use a Gated Recurrent Unit (GRU)~\cite{gru} as an RNN unit to illustrate the integration,  
where the matrix multiplications in classic GRU are replaced by the graph attention based diffusion convolutional process as defined in Equation~\ref{eq:diffusion}.   

\begingroup\abovedisplayskip=2pt
\begin{align*}
r_{t} &= \sigma(\mathbf{\Theta}_r \otimes [\mathbf{X}_t|| \mathbf{H}_{t-1}] + b_r) \\
u_{t} &= \sigma(\mathbf{\Theta}_u \otimes [\mathbf{X}_t|| \mathbf{H}_{t-1}] + b_u) \\
\hat{h}_{t} &= tanh(\mathbf{\Theta}_{\hat{h}} \otimes [\mathbf{X}_t||(r_{t} \odot \mathbf{H}_{t-1})] + b_{\hat{h}})\\
\mathbf{H}_{t} &= u_{t} \odot \mathbf{H}_{t-1} + (1-u_{t}) \odot \hat{h}_{t}
\end{align*}
\endgroup
Here, $\mathbf{X}_{t} \in \mathcal{R}^{N\times K}$ is the graph signal at timestamp $t$, $\mathbf{H}_{t}$ is the output at timestamp $t$. $r_{t}$, $u_{t}$ and $\hat{h}_{t}$ represent the reset gate, update gate, and candidate context at time $t$.  
$\otimes$ indicates the proposed graph attention based diffusion convolution, and $\mathbf{\Theta}_r$,  $\mathbf{\Theta}_u$, and $\mathbf{\Theta}_{\hat{h}}$ are the filters used in the three convolutions. 
%
$\odot$ is Hadamard product.

The proposed graph attention is generic in the sense that it provides a data-driven manner to produce dynamic adjacency matrices and thus can be integrated with different kinds of convolutions that utilize graph adjacency matrices. Such convolutions often use a static adjacency matrix, while the proposed graph attention allows us to employ dynamic adjacency matrices at different timestamps, which are expected to better capture the spatio-temporal correlations among different time series.  
%
%


\subsection{Loss Function}
The loss function measures the discrepancy between the estimated travel speed and the ground truth at each timestamp for each entity. Suppose that we have $Y$ training instances in total, we use Equation~\ref{eq:loss} to measure the discrepancy. 

\begin{equation}
    \mathit{loss}= \frac{1}{YNP} \sum_{y=1}^{Y} \sum_{n=1}^{N} \sum_{p=1}^{P} 
    |\mathbf{x}_{p,y}^{(n)} - \mathbf{\hat{x}}_{p,y}^{(n)}|
    \label{eq:loss}
\end{equation}
where $\mathbf{x}_{p,y}^{(n)}$ and $\mathbf{\hat{x}}_{p,y}^{(n)}$ are the ground truth and the estimated speed for the $n$-th entity at timestamp $p$ in the $y$-th training instance.

\section{Empirical Study}
\begin{figure*}[!ht]
\centering

\begin{tabular}{|c|l|ccc|ccc|ccc|}
\hline
\multicolumn{1}{|l|}{\multirow{2}{*}{Data}} & \multirow{2}{*}{Models} & \multicolumn{3}{c|}{15 min}                    & \multicolumn{3}{c|}{30 min}                    & \multicolumn{3}{c|}{60 min}                     \\ \cline{3-11} 
\multicolumn{1}{|l|}{}                      &                         & MAE           & RMSE          & MAPE           & MAE           & RMSE          & MAPE           & MAE           & RMSE          & MAPE            \\ \hline
\multirow{11}{*}{\rotatebox{90}{METR-LA}}                 & HA                      & 4.16          & 7.80          & 13.0\%         & 4.16          & 7.80          & 13.0\%         & 4.16          & 7.80          & 13.0\%          \\
                                            & ARIMA                   & 3.99          & 8.21          & 9.6\%          & 5.15          & 10.45         & 12.7\%         & 6.90          & 13.23         & 17.4\%          \\
                                            & VAR                     & 4.42          & 7.89          & 10.2\%         & 5.41          & 9.13          & 12.7\%         & 6.52          & 10.11         & 15.8\%          \\
                                            & SVR                     & 3.99          & 8.45          & 9.3\%          & 5.05          & 10.87         & 12.1\%         & 6.72          & 13.76         & 16.7\%          \\
                                            & FC-LSTM                 & 3.44          & 6.30          & 9.6\%          & 3.77          & 7.23          & 10.9\%         & 4.37          & 8.69          & 13.2\%          \\
                                            & WaveNet                 & 2.99          & 5.89          & 8.0\%          & 3.59          & 7.28          & 10.2\%         & 4.45          & 8.93          & 13.6\%          \\
                                            & STGCN                   & 2.88          & 5.74          & 7.6\%          & 3.47          & 7.24          & 9.5\%          & 4.59          & 9.40          & 12.7\%          \\ \cline{2-11} 
                                            & GCRNN                   & 2.80          & 5.51          & 7.5\%          & 3.24          & 6.74          & 9.0\%          & 3.70          & 8.16          & \underline{10.9}\%          \\
                                            & GA-GCRNN                & {\underline{2.76}}    & \underline{5.33}          & \underline{7.1}\%          & \underline{3.21}          & \underline{6.45}          & \underline{8.8}\%          & \underline{3.70}          & \underline{7.68}          & \underline{10.9}\%          \\ \cline{2-11} 
                                            & DCRNN                   & 2.77          & 5.38          & 7.3\%          & 3.15          & 6.45          & 8.8\%          & \textbf{3.60} & \textbf{7.60} & \textbf{10.5\%} \\
                                            & GA-DCRNN                & \textbf{2.75} & \textbf{5.28} & \textbf{7.0\%} & \textbf{3.14} & \textbf{6.39} & \textbf{8.0\%} & 3.72          & 7.64          & 11.0\%          \\ \hline
\end{tabular}

\captionof{table}{Accuracy at $p=3$, $p=6$, and $p=12$}
\vspace{-15pt}
\label{result_table}
\end{figure*}


\subsection{Experimental Setup}
We conducted experiments on a large real world traffic dataset METR-LA from \cite{DCRNN}. The dataset consists of speed measurements from 207 loop detectors spread across Los Angeles highways. The data was collected between March 1st 2012 and June 30th 2012 with a frequency of every 5 minutes. 

\par We follow the same experimental setup as~\cite{DCRNN}. 
We build a graph by connecting from sensors $i$ to $j$ if the road network distance from $i$ to $j$ is small~\cite{DCRNN}. 
Since road network distance is used, the distance from $i$ to $j$ may be different from the distance from $j$ to $i$, making the adjacency matrix $E$ asymmetric. 
We use 70\% of the data for training, 10\% for validation and 20\% for testing. We consider three metrics to evaluate the prediction accuracy: Mean Absolute Error (MAE), Root Mean Square Error (RMSE), Mean Absolute Percentage Error (MAPE). We consider the same forecasting setting as~\cite{DCRNN} where we use $l=12$ past observations to predict the next $p=12$ steps ahead and report the errors at three different intervals (15, 30, and 60 mins) and also the average errors over $p=12$ steps. 

\subsection{Implementation Details} The method is implemented in Python 3.6 using Tensorflow 1.7. A server with Intel Xeon Platinum 8168 CPU and 2 Tesla V100 GPUS are used to conduct all experiments. We trained the model using Adam optimizer with 0.01 as learning rate which decreases every 10 epochs after the 40th iteration. We used a total of $C=2$ attention heads with an embedding size of $F=16$. In addition we used a 2 layer GRU, with 64 units, a batch size of 64, and the scheduled sampling technique as described in \cite{DCRNN}.

\noindent
\subsection{Experimental Results}
\noindent
\textbf{Baselines: } We compared our proposed model with the following methods:
\begin{itemize} 
\setlength\itemsep{0.01em}
    \item \texttt{HA} historical average~\cite{DCRNN}. \texttt{HA}  models the data as a seasonal process and computes the predictions as a weighted average of the previous seasons.
    \item \texttt{ARIMA} \cite{arima} integrated moving average model with Kalman filter which is widely used for time series forecasting.
    \item \texttt{VAR}~\cite{VAR} vector auto-regression.  
    \item \texttt{SVR} support vector regression.
    \item \texttt{FC-LSTM} \cite{FCLSTM} recurrent neural network with fully connected LSTM units. 
    \item \texttt{WaveNet} \cite{WaveNet} a dilated causal convolution network. 
    \item \texttt{STGCN} \cite{STGCN} spatial-temporal  GCN  that combines 1D convolution with graph convolution. 
    \item \texttt{GCRNN} \cite{DCRNN} graph convolutional recurrent neural network. 
    \item \texttt{DCRNN} \cite{DCRNN} diffusion convolutional recurrent neural network. 
\end{itemize}
A more in-dept description of the baselines, along with hyper-parameters can be found in~\cite{DCRNN}. 
Since \texttt{GCRNN} and \texttt{DCRNN} employ graph based convolutions, where \texttt{GCRNN} uses graph convolution and \texttt{DCRNN} uses diffusion convolution~\cite{DCRNN}, we incorporate the proposed graph attention based adaptive adjacency matrix into the two methods to obtain \texttt{GA-GCRNN} and \texttt{GA-DCRNN}. 

\noindent
\textbf{Accuracy: } 
We compare the proposed \texttt{GA-GCRNN} and \texttt{GA-DCRNN} with the baselines in Table~\ref{result_table}. 
Firstly, we observe that non-deep learning methods such as \texttt{ARIMA}, \texttt{VAR} and \texttt{SVR} perform poorly compared to the deep learning methods especially when  $p$ is large. This is because the temporal dependencies becomes increasingly non-linear as the horizon increases and the aforementioned baselines are unable to capture such dynamics.    

We also observe that by considering the underlying correlations among entities using a graph is very effective on improving the accuracy. Here, FC-LSTM and WaveNet do not consider such correlations and are outperformed by the other deep learning methods that consider the correlations. 

%

Next, we evaluate the proposed graph attention method for generating the dynamic adjacency matrices. 
We first compare \texttt{GA-GCRNN} with \texttt{GCRNN}, where \texttt{GA-GCRNN} outperforms \texttt{GCRNN} in all settings (see the underline values in Table~\ref{result_table}).  
We then consider GA-DCRNN. \texttt{GA-DCRNN} outperforms \texttt{DCRNN} in most settings (see the bold values in Table~\ref{result_table}), especially in short term predictions at 15 and 30 minutes. 

We also consider the average accuracy over all the 12 steps in the prediction, as shown in Table~\ref{tbl:avg}.   
On average, \texttt{GA-DCRNN} is better when using RMSE and MAPE. This suggests that \texttt{GA-DCRNN} avoids having large prediction errors but may have more small prediction errors than does \texttt{DCRNN}. This shows that \texttt{GA-DCRNN} better captures the general trend of the underlying traffic data.   

\begin{table}[!h]
\centering
\begin{tabular}{|l|ccc|}
\hline
\multirow{2}{*}{Models} & \multicolumn{3}{c|}{Average}                    \\ \cline{2-4} 
                        & MAE           & RMSE          & MAPE            \\ \hline
GCRNN                   & 3.28          & 6.80          & 9.13\%          \\
GA-GCRNN                & \underline{3.22}          & \underline{6.48}          & \underline{8.93}\%          \\ \hline
DCRNN                   & \textbf{3.17} & 6.47          & 8.86\%          \\
GA-DCRNN                & 3.22          & \textbf{6.43} & \textbf{8.66\%} \\ \hline
\end{tabular}
\captionof{table}{Average Accuracy}
\label{tbl:avg}
\vspace{-10pt}
\end{table}

The proposed graph attention based method for generating dynamic adjacency matrices is effective, which helps improve prediction accuracy, especially for short term prediction. In addition, the proposed method is generic as it can be used to boost the accuracy of both DCRNN and GCRNN. 

\noindent
\textbf{Efficiency: }While \texttt{DCRNN} takes on average 271 seconds for one epoch, \texttt{GA-DCRNN} requires 401 seconds. The discrepancy between the two models is due to the attention mechanism, which requires more computation. Note that when implementing the attention we followed~\cite{graph_attention}, but a more efficient way is available~\cite{geniepath}. For the same reason, \texttt{GA-GCRNN} also takes longer time than \texttt{GCRNN} does. 

\noindent
\textbf{Summary: }
The results suggest that the proposed graph head attention based adaptive adjacency matrices can be easily integrated with convolutions that consider graph topology and has a great potential to enable more accurate predictions, especially for relatively short term predictions. It is of interest to further investigate how to improve long term predictions. 
%

%

\section{Related work}

\par 
For time series forcasting, auto-regressive models, e.g., ARIMA, is widely used as a baseline method. 
%
Hidden Markov models are also often used to enable time series forecasting. A so-called spatio-temporal HMM (STHMM) is able to consider spatio-temporal correlations among traffic time series from adjacent edges~\cite{DBLP:journals/pvldb/0002GJ13}. 
%

Neural networks are able to capture non-linear dependencies within the data, which enable non-linear forecasting models and this category of models started to receive greater attention in the last years. 
In particular, Recurrent Neural Networks (RNNs) are used with successes in multiple domains such as traffic time series prediction~\cite{DCRNN} or wind forecasting \cite{wind_forecasting}, due to their recurrent nature that is able to capture temporal relationships within the data.  However they suffer from the well known vanishing gradient problem \cite{vanishing_problem}. Different models such as Long-Short-Term-Memory (LSTM) or Gated-Recurrent-Unit (GRU) come as an extension to traditional RNNs. The core idea of RNN models is by integrating different gates to control what the network is going to learn and forget leading to better results  and an elegant solution for the vanishing problem. 
%

Another well known type of deep learning models are Convolutions Neural Networks (CNNs). CNNs are considered to be the state of the art when it comes to speech or image recognition. CNN based models are well known for their ability to capture spatial relationships, nearby elements, e.g., pixels in images, share local features by traversing the input multiple times and applying different filters. CNNs often work on grid-based inputs, limiting its applicability to graph-based inputs such as road networks. Recently, Bruna et al. \cite{gcn_first} introduce Graph Convolutional Networks (GCNs) that combines spectral graphs theory with CNNs. Two recent studies employ GCNs to fill in missing values for uncertain traffic time series~\cite{icde2019,icde20hu}. 
\cite{diffusion_convolution} proposes Diffusion Convolution Neural Networks (DCNNs) which fall under the non-spectral methods. DCNN works on the assumption that the further two nodes are in terms of graph topology, the lower impact they should have. 

\par DCRNN~\cite{DCRNN} extends DCNN with RNN so that it also captures time dependencies within the data. 
Our work is closely related and builds on top of DCRNN. The main difference is that DCRNN assumes that the adjacency matrix used in random walks is static. However, this assumption might not always hold and
we propose to learn an adaptive adjacency matrix at each time stamp using attention mechanism that considers graph topology. 
Multi-task learning~\cite{DBLP:conf/cikm/Kieu0GJ18} has also been applied for correlated time series forecasting, where both a CNN and a RNN are combined to both forecast future values and reconstruct historical values~\cite{DBLP:conf/cikm/CirsteaMMG018}. %
Our work resembles \cite{graph_attention} and \cite{gaan}. The main difference is that our model tries to learn a dynamic adjacency matrix which can afterwards be used with any type of graph-RNN like structure to update each node embedding by taking into account its neighbours offering more flexibility.

From application perspectives, many smart transportation applications~\cite{simonicde,cguo,l2rsean,DBLP:journals/pvldb/PedersenYJ20,DBLP:conf/gis/AljubayrinYJZ16,simon} rely on accurate travel speed estimation, which can benefit from our work. 
\section{Conclusion and Outlook}

We propose a generic method to obtain dynamic adjacency matrices using graph attention which can be integrated seamlessly with existing graph based convolutions such as graph convolution and diffusion convolution. More specifically, we show how the integration of adaptive adjacency matrices and recurrent neural networks is able to improve the correlated time series predictions where the relationships among different time series can be captured as a graph. Experimental results show great potential when using adaptive adjacency matrices, especially for short term predictions.  

As future work, it is of interest to explore how speed up the attention computation using, e.g., parallel computation~\cite{DBLP:conf/waim/YuanSWYZY10}. It is also of interest to explore how to incorporate additional knowledge of the entities, e.g., points of interest around the speed sensors.  

\balance
\bibliographystyle{IEEEtran}
\bibliography{references}
\end{document}